\documentclass[letterpaper, 10pt, journal]{ieeeconf}  % Comment this line out if you need a4paper

\IEEEoverridecommandlockouts                              % This command is only needed if 
\usepackage{amsmath} % assumes amsmath package installed
\usepackage{amsfonts}
\usepackage{pgf}
\usepackage{hyperref}
\usepackage{mathtools}
\usepackage{graphicx}
\usepackage{import}
\usepackage{xcolor}
\usepackage{graphicx}

\makeatletter
\newcommand*\titleheader[1]{\gdef\@titleheader{#1}}
\AtBeginDocument{%
  \let\st@red@title\@title
  \def\@title{%
    \bgroup\normalfont\large\centering\@titleheader\par\egroup
    \vskip1.5em\st@red@title}
}
\makeatother

\title{\LARGE \bf
Reinforcement Learning for Robust Athletic Intelligence: Lessons from the 2nd
“AI Olympics with RealAIGym” Competition
}
\titleheader{\vspace{-1cm}\small{This work has been submitted to the IEEE for
possible publication. Copyright may be transferred without notice, after which
this version may no longer be accessible.}}

\author{Felix Wiebe$^{1}$, 
Niccolò Turcato$^{2}$,
Alberto Dalla Libera$^{2}$,
Jean Seong Bjorn Choe$^{3}$,
Bumkyu Choi$^{3}$,\\
Tim Lukas Faust$^{2}$, 
Habib Maraqten$^{2}$,
Erfan Aghadavoodi$^{1}$, 
Marco Cali$^{2}$, 
Alberto Sinigaglia$^{2}$,\\
Giulio Giacomuzzo$^{2}$,
Diego Romeres$^{5}$,
Jong-kook Kim$^{3}$,
Gian Antonio Susto$^{2}$,\\
Shubham Vyas$^{3}$,
Dennis Mronga$^{3}$, 
Boris Belousov$^{4}$,
Jan Peters$^{4,6,9,10}$,
Frank Kirchner$^{1,7}$
and Shivesh Kumar$^{1,8}$% <-this % stops a space
\thanks{$^{1}$Robotics Innovation Center, German Research Center for Artificial
Intelligence (DFKI), Germany}%
\thanks{$^{2}$Department of Information Engineering, University of Padova, Italy}%
\thanks{$^{3}$Korea University, Seoul, South Korea}%
\thanks{$^{4}$Systems AI for Robot Learning, German Research Center for Artificial Intelligence (DFKI), Germany}%
\thanks{$^{5}$Mitsubishi Electric Research Lab (MERL), USA}%
\thanks{$^{6}$Technical University of Darmstadt, Germany}%
\thanks{$^{7}$University of Bremen, Germany}%
\thanks{$^{8}$Chalmers University of Technology, Sweden}%
\thanks{$^9$Center for Cognitive Science, Germany}\\
\thanks{$^{10}$Hessian.AI, Germany}
}

\begin{document}

\maketitle
\thispagestyle{empty}
\pagestyle{empty}

%%%%%%%%%%%%%%%%%%%%%%%%%%%%%%%%%%%%%%%%%%%%%%%%%%%%%%%%%%%%%%%%%%%%%%%%%%%%%%%%
\begin{abstract}

In the field of robotics many different approaches ranging from classical
planning over optimal control to reinforcement learning (RL) are developed and
borrowed from other fields to achieve reliable control in diverse tasks. In
order to get a clear understanding of their individual strengths and weaknesses
and their applicability in real world robotic scenarios is it important to
benchmark and compare their performances not only in a simulation but also on
real hardware. The '2nd AI Olympics with RealAIGym' competition was held at the
IROS 2024 conference to contribute to this cause and evaluate different
controllers according to their ability to solve a dynamic control problem on an
underactuated double pendulum system (Fig. \ref{fig:onsite}) with chaotic
dynamics. This paper describes the four different RL methods submitted by the
participating teams, presents their performance in the swing-up task on a real
double pendulum, measured against various criteria, and discusses their
transferability from simulation to real hardware and their robustness to
external disturbances.
\end{abstract}

%%%%%%%%%%%%%%%%%%%%%%%%%%%%%%%%%%%%%%%%%%%%%%%%%%%%%%%%%%%%%%%%%%%%%%%%%%%%%%%%
\section{Introduction}

% general introduction (abstract)

\begin{figure}[t]
%\label{fig:onsite}
\centering
\includegraphics[width=\columnwidth]{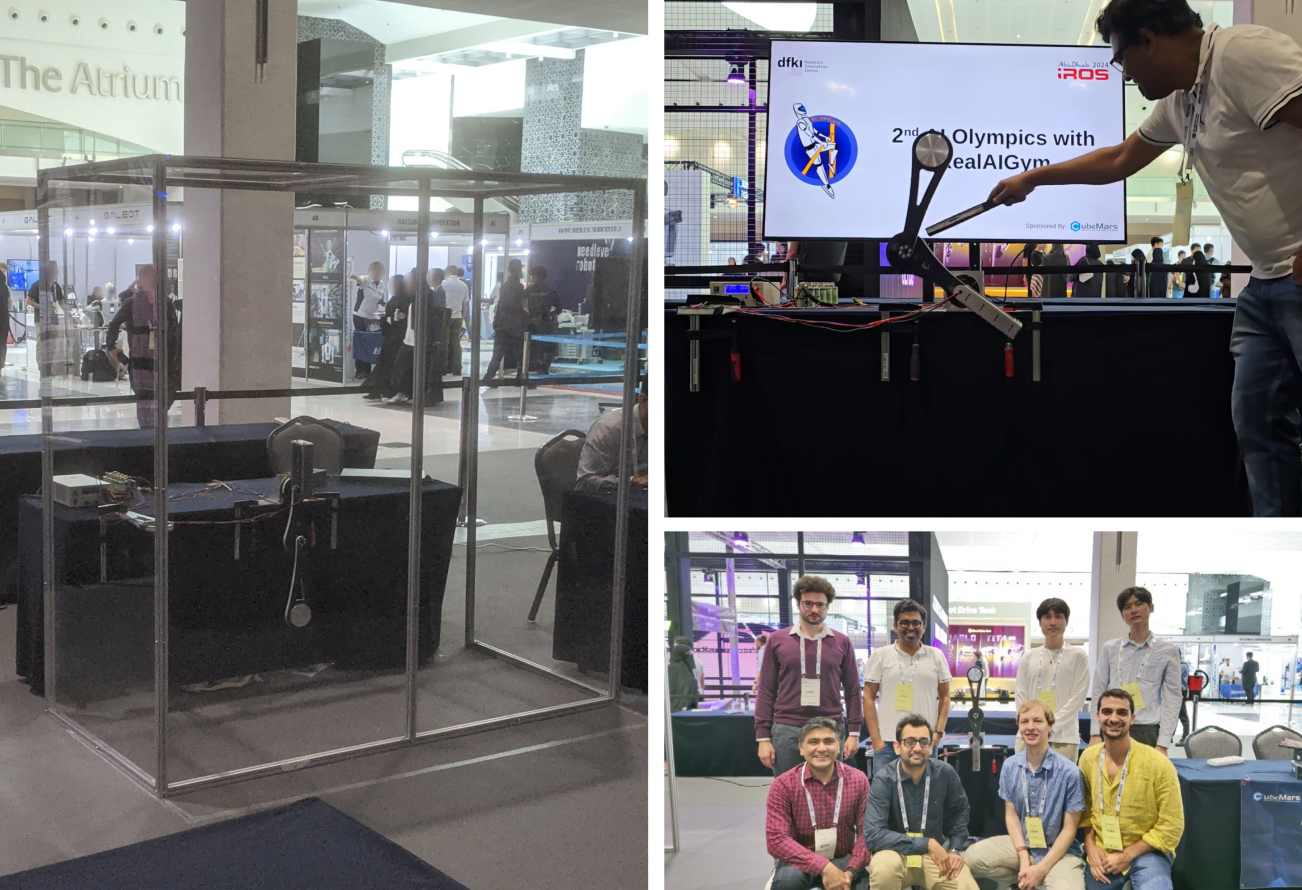}
\caption{Onsite setup of the double pendulum hardware (left), the pendulum being
disturbed with a stick (top right) and participating teams (bottom right) at
IROS 2024.}
\label{fig:onsite}
\end{figure}

% related work
There are many frameworks for comparing control methods in simulation
environments (e.g., \cite{c21, james2020rlbench}) but only few provide a
standardized real robot environment which is open and easily reproducible such
as \cite{grimminger2020open} with a quadruped robot and
\cite{funk2022benchmarking} with a three end-effector robot for manipulation
tasks. The RealAIGym project focuses on simple, low degree of freedom systems
capable of dynamic behaviors, such as a simple pendulum \cite{Wiebe2022torque},
AcroMonk \cite{javadi2023acromonk}, RicMonk~\cite{2024_icra_ricmonk} and the
dual purpose double pendulum \cite{wiebe2024doublependulum} used in this
competition. With the software and hardware being open source and a user
friendly Python API, RealAIGym intends to be an open benchmarking platform for
comparing control methods on real hardware. This is the 2nd iteration of this
competition after the first one having been held at the IJCAI 2023 conference
\cite{wiebe2024competition1}.

\section{Competition Rules}
The competition was conducted with the dual purpose double pendulum system
introduced in \cite{wiebe2024doublependulum} which can be operated as an acrobot
or pendubot without changing the hardware. For the competition acrobot and
pendubot were treated as separate challenges and evaluated in separate tracks.
The task on both systems is to swing up the pendulum from the free hanging
position to the upright position and stabilize it there. To test the
controllers' robustness, external perturbations are applied on both motors
during the execution. The used system had an attached mass of $0.5\,\text{kg}$
and link lengths of $0.2\text{m}$ and $0.3\text{m}$. Model parameters identified
with least squares optimization were provided to the participants, but the teams
were free to do their own system identification.

The competition was carried out in three phases: (i) A simulation phase, (ii) a
remote hardware phase and (iii) an onsite phase. In the simulation phase the
participants were asked to develop a controller for the swing-up and balance
task on the acrobot and/or pendubot system and integrate it into the double
pendulum toolkit on
github\footnote{\href{https://github.com/dfki-ric-underactuated-lab/double_pendulum}{https://github.com/dfki-ric-underactuated-lab/double\_pendulum}}.
After the simulation phase, the four best performing controllers were advanced
to the next phase. In the remote hardware phase, the teams got the opportunity
to remotely operate the pendulum hardware and collect data from the real system
along with live video feed. The third phase took place on site at the IROS 2024
conference, where the teams were able to fine-tune their controllers before the
final evaluation.

For the evaluation of the controllers different criteria were considered.
Simulation results were evaluated with the \textit{performance score}, where the
controllers show their basic functionality. and with the \textit{robustness
score}, where the controllers show their robustness to external influences.

The formula for the performance score is:
\begin{align}
    S_{p} = c_{succ} \left[ 1 - \frac{1}{5} \sum_{i \in \{ t, E, \tau_c, \tau_s, v\}} \tanh{(w_{i}c_i)}\right]
    \label{eq:perf_score}
\end{align}
based on the swing-up success $c_{succ} \in \{ 0, 1\}$, the swing-up time $c_t$,
the used energy $c_E$, the torque cost $c_{\tau_c}$, the torque smoothness
$c_{\tau_s}$ and the velocity cost $c_v$. These quantities are normalized with
the weights in Table \ref{tab:perf_score_weights}
and scaled with the $\tanh$ function to the interval $[0, 1]$. 

The second score computed from the simulation results is the robustness score:
\begin{align}
    S_{r} = \frac{1}{6} \big( c_{m} + c_{v, \text{noise}} + c_{\tau,
    \text{noise}} + c_{\tau, \text{resp.}} + c_{d} + c_{p} \big)
\end{align}
with the criteria
\begin{itemize}
\item Model inaccuracies $c_m$: The independent model parameters are varied one
at the time in the simulator while using the original model parameters in the
controller.
\item Measurement noise $c_{v, \text{noise}}$: Artificial noise is added to the
velocity measurements.
\item Torque noise $c_{\tau, \text{noise}}$: Artificial noise is added to the
motor torque.
\item Torque response $c_{\tau, \text{resp.}}$: A delayed motor response is
modeled by applying the torque $\tau = \tau_{t - 1} + k_{resp}(\tau_{des} -
\tau_{t - 1})$ instead of the desired torque $\tau_{des}$, where, $\tau_{t-1}$
is the applied motor torque from the last time step and $k_{resp}$ is the factor
which scales the responsiveness.
\item Time delay $c_{d}$: A time delay is added to all measurements.
\item Random perturbations $c_{p}$: A randomly generated sequence of Gaussian
perturbations is added to the applied torque.
\end{itemize}
For each criterion the severity of the external influence is varied in $N=21$
steps (for the model inaccuracies for each independent model parameter) and the
score is the percentage of successful swing-ups under these circumstances. For
the perturbation criterion 50 random perturbations profiles are generated and
evaluated.

For hardware phases, only the performance score, $S_p$ (\ref{eq:perf_score}), was
calculated with slightly adjusted weights (Table
\ref{tab:perf_score_weights}), but the controllers had to showcase their
robustness by reacting to unknown external perturbations during the experiment.
The perturbations consist of Gaussian torque profiles on both joints at random
times, which were unknown to the controllers. Each trial lasts for
$10\,\text{s}$ and is considered successful if, in the end, the end-effector is
at a height above the threshold of $0.45\,\text{cm}$ above the base. The teams
were allowed to use up to $0.5\,\text{Nm}$ torque on the passive joint for
friction compensation. Each controller was tested in 10 trials and the final
score is the average of the individual scores.

\begin{table}[ht]
    \centering
    \caption{Weights for the performance score (\ref{eq:perf_score}).}
    \label{tab:perf_score_weights}
    \begin{tabular}{cccccc}
                    & $w_t$         & $w_E$         & $w_{\tau_c}$  & $w_{\tau_s}$  & $w_v$     \\
        \hline
        Simulation  & $\pi / 20$    & $\pi / 60$    & $\pi/20$      & $10\pi$       & $\pi/400$ \\
        Hardware    & $\pi / 20$    & $\pi / 60$    & $\pi/100$     & $\pi/ 4.0$    & $\pi/400$ \\

    \end{tabular}
\end{table}

\section{Algorithms}

Four teams advanced to the final round of the competition and tested their
control methods on the real hardware. This section briefly introduces the four
methods and how they approached the task to improve robustness.

\subsection{Model-based RL: MC-PILCO}
\begin{figure*}
    \centering
    \includegraphics[width=\linewidth]{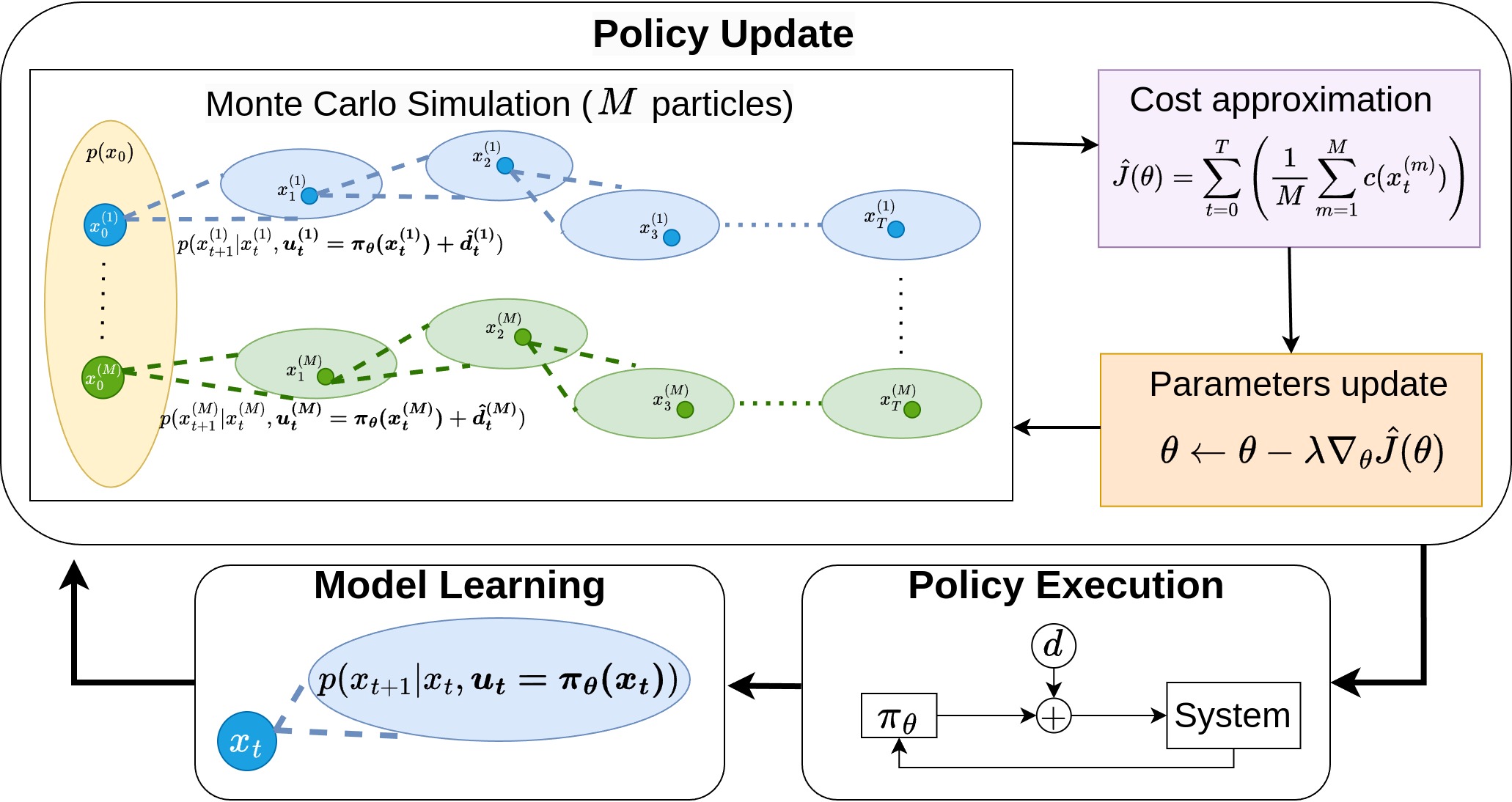}
    \caption{Schematic representation of the robust MC-PILCO algorithm. Each
    Episode is composed of (i) model learning, (ii) policy update and
    (iii) policy execution.}
    \label{fig:mc_pilco}
\end{figure*}
MC-PILCO (Monte Carlo - Probabilistic Inference for Learning COntrol)
\cite{amadio2022mcpilco} is a Model-Based policy gradient RL algorithm. MC-PILCO
relies on Gaussian Processes (GPs) to learn the system dynamics from data, and a
particle-based policy gradient optimization procedure to update the policy
parameters. Let $\boldsymbol{x}_{t}$ and $\boldsymbol{u}_{t}$ be, respectively,
the state and input of the system at step $t$. A cost function
$c(\boldsymbol{x}_{t})$ encodes the task to solve. A policy
$\pi_{\boldsymbol{\theta}}: \boldsymbol{x} \rightarrow \boldsymbol{u}$
parametrized by $\boldsymbol{\theta}$ selects the inputs applied to the system.
The objective is to find the policy parameters $\boldsymbol{\theta}^*$ which
minimize the cumulative expected cost:
\begin{equation}
    J(\boldsymbol{\theta}) = \sum_{t=0}^T \mathbb{E} [c(\boldsymbol{x}_{t})] \hspace{0.2cm} x_0 \sim p(\boldsymbol{x}_0)
    \label{eq:mc_pilco_cumulative_cost},
\end{equation}
where $c(\boldsymbol{x}_{t})$ is a saturated distance as used in
\cite{amadio2022mcpilco} and \cite{turcato2023teaching}.

MC-PILCO iterates attempts to solve the objective task, also called trials. 
Each trial is composed of three main steps: (i) model learning, (ii) policy
update, and (iii) policy execution. Fig. \ref{fig:mc_pilco} illustrates the main
details of each phase.

In the model learning step, previous experience is used to derive a
one-step-ahead stochastic system dynamics model, using Gaussian Process
Regression. The policy update step aims at minimizing the cost in
(\ref{eq:mc_pilco_cumulative_cost}) w.r.t. $\boldsymbol{\theta}$. The expectation
in (\ref{eq:mc_pilco_cumulative_cost}) is approximated using the GP dynamics
derived in step (i) and Monte Carlo particle-based simulation. Finally, in the
last step, the new optimized policy is executed on the system and the collected
data is stored to update the model in the next trials. 
%Examples of MC-PILCO applications have been reported in \cite{amadio2023mcpilco_raw_meas} and \cite{mcpilco_tossing}.

Generally, a well-trained policy can recover if disturbances push the system
away from its trajectory. For the sake of the competition, it is preferred to
train a policy that can counter these disturbances, without falling and
re-maneuvering. To make the policy more robust to these perturbations, we
simulate their effects during policy optimization. Namely, for each particle we
compute a disturbance profile $\boldsymbol{\hat{d}}^{(m)}_t$, $m = 1, \dots, M$,
$t=0,\dots,T$, using the known disturbance profile distribution.
Then, when performing the estimation of the forward dynamics in the policy
optimization step, the one-step-ahead evolution of the particles is computed as 
\begin{equation}
    \boldsymbol{x}_{t+1}^{(m)} = \boldsymbol{x}_{t}^{(m)} + \hat{f}(\boldsymbol{x}_t, \pi_\theta(\boldsymbol{x}_t^{(m)}) + \boldsymbol{\hat{d}}_t^{(m)}),
\end{equation}
where $\hat{f}(\cdot, \cdot)$ is the forward dynamics model estimated in the
model learning phase.
The cost function encourages the policy to reject such disturbances while
limiting additional maneuvers.

\subsection{Model-free RL, Average-Reward Entropy Advantage Policy Optimization
(AR-EAPO)}

Average-Reward Entropy Advantage Policy Optimization (AR-EAPO) is a model-free
RL algorithm that extends Entropy Advantage Policy Optimization (EAPO)
\cite{choe2024maximumentropyonpolicyactorcritic}, a maximum entropy (MaxEnt)
on-policy actor-critic algorithm, into the average-reward setting.

AR-EAPO tackles the challenge of applying model-free RL to this complex problem
by formulating the system as a recurrent Markov Decision Process (MDP), where
every state is reachable from every other state in a finite number of steps
\cite{puterman2014markov}. With the formulation, instead of using discounted
cumulative rewards,  AR-EAPO optimizes for long-term average rewards. This
approach enables the use of straightforward reward functions aiming for
long-term optimality rather than sophisticated reward functions for discounted
settings. Here, the proposed solution utilizes a simple quadratic cost function.

Specifically, the objective of AR-EAPO is to optimize the \textit{gain} – the
sum of expected average reward and entropy $\tilde{\rho}^\pi$ for a stationary
policy $\pi$:

{
\small
\begin{align}
    \tilde{\rho}^\pi(s) \coloneqq \left. \lim_{T\rightarrow\infty}\frac{1}{T}\mathbb{E}\left[
        \sum^{T-1}_{t=0}r_t-\tau\log\pi(a_t|s_t) \right| s_0=s,s_t\sim\pi
    \right],
\end{align}
}
where $s_t$ is the state of the MDP at time $t$, $a_t$ is the action at time
$t$, and $r_t$ is the reward received at time $t$. Building on EAPO's framework,
AR-EAPO separately estimates reward and entropy objectives.

Additionally, it incorporates a gain approximation mechanism using advantage
estimations \cite{dewanto2020average}:
\begin{align}
    \tilde{\rho}^{k+1} = \tilde{\rho}^k + \eta\mathbb{E}_t[\tilde{A}^\pi(s_t,a_t)],
\end{align}
where $\tilde{\rho}^{k}$ represents the gain estimate at $k$-th iteration,
$\tilde{A}^\pi$ denotes the soft advantage function, $\eta$ is the step size
hyperparameter. The algorithm samples state-action pairs $(s_t, a_t)$ from
rollout trajectories to estimate the advantages using generalized advantage
estimation (GAE) \cite{schulman2015high}. Thus, this formulation effectively
extends the well-established proximal policy optimization  (PPO)
\cite{schulman2017proximal} to the average-reward MaxEnt RL setting.

\begin{figure}[t]
    \centering
    \includegraphics[width=\columnwidth]{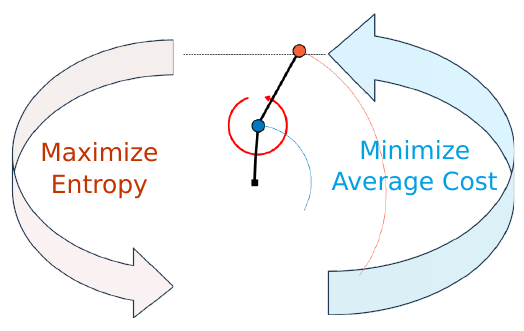}
    %\import{images}{AR-EAPO_overview_new.pdf_tex}
    \caption{Illustration of AR-EAPO in the double pendulum system: learning
    efficient and robust control in an infinite loop of swing-ups while
    balancing the cost and the entropy.}
\label{fig:AR-EAPO}
\end{figure}

In the double pendulum system, AR-EAPO demonstrates an interesting learning
pattern. While the cost objective drives the controller to develop an efficient
swing-up policy, the entropy component prevents the pendulum from remaining
stationary at the uppermost position (Fig. \ref{fig:AR-EAPO}).
Instead, it encourages movement toward lower positions where average entropy is
higher. Therefore, through these endless swing-ups and downs, the average-reward
MaxEnt RL formulation naturally promotes diverse trajectories, thereby achieving
a robust control policy.

\subsection{Model-free RL, Actor-Critic Method: EvolSAC}
\label{sec:evolsac}
This section presents the EvolSAC algorithm for the AI Olympics competition. The
approach integrates model-free deep RL with evolutionary strategies to optimize
control performance in both simulation and real hardware stages. EvolSAC is
presented in more detail in \cite{cali2024EVOSAC}.

The core algorithm used is Soft Actor-Critic (SAC) \cite{haarnoja2018SAC}, a
state-of-the-art RL method designed for continuous action spaces. SAC is a
model-free RL algorithm that optimizes a stochastic policy by maximizing both
the expected reward and the entropy of the policy, which promotes exploration
and robustness. SAC's objective function is defined as:
\[
J(\pi) = \mathbb{E}_{s_t, a_t \sim \pi} \left[ \sum_{t} \gamma^t \left( r(s_t,
a_t) + \alpha \mathcal{H}(\pi(\cdot | s_t)) \right) \right],
\]
where \( \gamma \) is the discount factor, \( r(s_t, a_t) \) is the reward
function, \( \alpha \) is a temperature parameter that weighs the importance of
the entropy term \( \mathcal{H} \), encouraging the policy to explore a wide
range of actions.

The initial stage of the methodology involves training the SAC agent with a
physics-inspired surrogate reward function. This function is designed to
approximate the complex objectives of the competition, which includes factors
like swing-up success, energy consumption, and torque smoothness. The surrogate
reward function aims to guide the agent towards achieving the swing-up task
while managing the energy costs effectively:
\begin{align}
R(s,a) = \begin{cases}
V + \alpha [1+ \cos(\theta_2)]^2 - \beta T & \text{ if } y > y_{th}\\
\,\,\,\,\,\,\,\,\,\,\,\,\,\,\,\,\,\,\,\,\,\,\,\,\,\,\,\,\,\,-\rho_1 a^2 - \phi_1 \Delta a\\
V - \rho_2 a^2 - \phi_2 \Delta a -\eta ||\dot{s}||^2  & \text{ otherwise }
\end{cases}
\end{align}
where $\alpha, \beta, \phi, \eta, \rho$ are hyperparameters used to control the
trade-offs.

To address the challenge of optimizing a policy for both performance and
robustness, the methodology incorporates evolutionary strategies in the later
stages of training. Evolutionary strategies are gradient-free optimization
methods that are particularly effective in scenarios where the landscape of the
objective function is rugged or noisy. One specific strategy used is the
Separable Natural Evolution Strategy (SNES) \cite{snes}, which updates mutation
strengths using a log-normal distribution, allowing efficient exploration of the
parameter space:
\begin{align}
    \begin{bmatrix}
     \sigma_{\text{new},i}\\
     \theta_{\text{new},i}
    \end{bmatrix}
    =
    \begin{bmatrix}
        \sigma_{\text{old},i} \exp \left( \tau \mathcal{N}(0, 1) + \tau'
        \mathcal{N}(0, 1)_i \right)\\
        \theta_{\text{old},i} + \sigma_{\text{new},i} \mathcal{N}(0, 1)_i
    \end{bmatrix}
\end{align}
where \( \tau, \tau' \) are learning rates controlling the mutation rate.

\begin{figure}[t]
    \centering
    %\includegraphics[width=\columnwidth]{images/competition evolsac.pdf}
    %\input{images/competition evolsac.pdf_tex}
    %% Creator: Inkscape 1.4 (1:1.4+202410161351+e7c3feb100), www.inkscape.org
%% Accompanies image file 'competition evolsac.pdf' (pdf, eps, ps)
\begingroup%
  \makeatletter%
  \providecommand\color[2][]{%
    \errmessage{(Inkscape) Color is used for the text in Inkscape, but the package 'color.sty' is not loaded}%
    \renewcommand\color[2][]{}%
  }%
  \providecommand\transparent[1]{%
    \errmessage{(Inkscape) Transparency is used (non-zero) for the text in Inkscape, but the package 'transparent.sty' is not loaded}%
    \renewcommand\transparent[1]{}%
  }%
  \providecommand\rotatebox[2]{#2}%
  \newcommand*\fsize{\dimexpr\f@size pt\relax}%
  \newcommand*\lineheight[1]{\fontsize{\fsize}{#1\fsize}\selectfont}%
  \ifx\svgwidth\undefined%
    \setlength{\unitlength}{251.10720634bp}%
    \ifx\svgscale\undefined%
      \relax%
    \else%
      \setlength{\unitlength}{\unitlength * \real{\svgscale}}%
    \fi%
  \else%
    \setlength{\unitlength}{\svgwidth}%
  \fi%
  \global\let\svgwidth\undefined%
  \global\let\svgscale\undefined%
  \makeatother%
  \begin{picture}(1,0.62348877)%
    \lineheight{1}%
    \setlength\tabcolsep{0pt}%
    \put(0,0){\includegraphics[width=\unitlength,page=1]{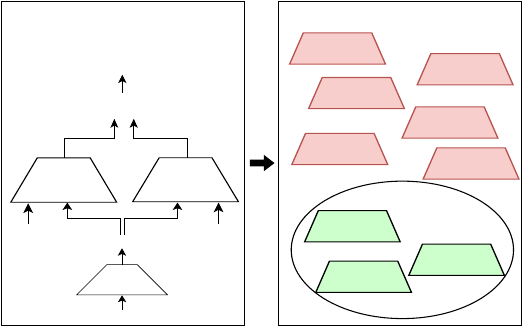}}%
    \put(0.14328975,0.58395121){\makebox(0,0)[lt]{\lineheight{1.25}\smash{\begin{tabular}[t]{l}SAC Training\end{tabular}}}}%
    \put(0.60805636,0.58395121){\makebox(0,0)[lt]{\lineheight{1.25}\smash{\begin{tabular}[t]{l}Evolutionary Training\end{tabular}}}}%
    \put(0.76769959,0.21475132){\makebox(0,0)[lt]{\lineheight{1.25}\smash{\begin{tabular}[t]{l}Preserve\\only top \textit{K}\end{tabular}}}}%
    \put(0,0){\includegraphics[width=\unitlength,page=2]{competition_evolsac.pdf}}%
  \end{picture}%
\endgroup%

    \caption{Summary of training for EvolSAC. Left: SAC training for optimal
    policy, right: Evolutionary selection of high scoring and robust policies.}
    \label{fig:evolsac}
\end{figure}

SNES is used to fine-tune the policy obtained from SAC, directly optimizing
against the competition’s score function defined in (\ref{eq:perf_score}). This
two-step process, initial training with SAC followed by fine-tuning with
SNES, ensures that the developed controller is not only effective in achieving
the desired task but also robust against various disturbances and model
inaccuracies. The overall training strategy is illustrated in
Fig. \ref{fig:evolsac}.

%\subsubsection{Velocity-History-Based Soft Actor-Critic}
\subsection{Model-free RL, Actor-Critic Method: HistorySAC}
\begin{figure}[t]
    \centering
    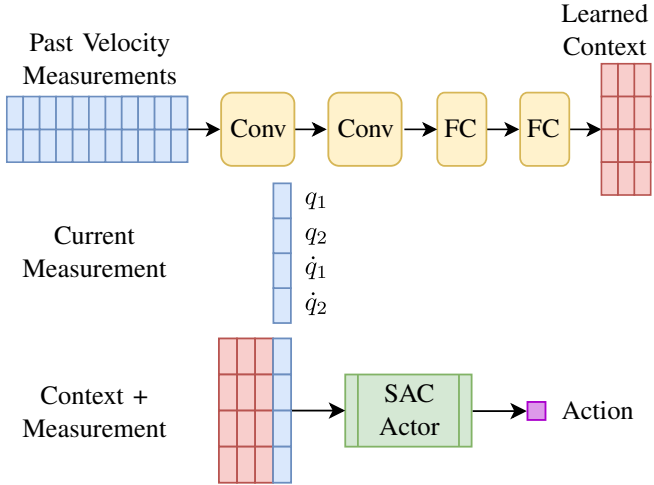
    \caption{The model architecture for encoding the history into a context
    representation in HistorySAC. A sequence of past velocity measurements is
    passed through convolutional (Conv) and fully-connected (FC) layers, and the
    output is attached to the current measurement before being passed to the
    actor and critic in SAC.}
    \label{fig:historysac_arch}
\end{figure}

This section proposes a solution for the challenge that combines the Soft
Actor-Critic algorithm with a method for learning temporal features. The
intuition behind this method is that it is not possible to infer system
parameters, such as the masses of both links, from a single state. This method
aims to implicitly encode the system dynamics by extracting temporal features
with convolutional and linear layers from previous measurements. For this, the
velocities of eleven previous and the current time step are encoded in a learned
temporal context and this context is concatenated with the current state before
feeding it into the actor and critic networks (Fig. \ref{fig:historysac_arch}).
The temporal information is encoded by using two 1-dimensional convolutional
layers with a kernel size of five and an output size of twelve, followed by two
linear layers with a width of 256, where the first uses a ReLU and the second
uses a tanh activation function. The underlying reinforcement learning
algorithm is the Stable-Baselines3 implementation of SAC~\cite{c21}, but with a
layer size of $1024$ instead of the original $256$ for the fully-connected
layers. It is important to note that the networks of actor and critic do not
share parameters. 

The training uses the reward function
\begin{equation}
\label{reward_history_sac}
\begin{aligned}
R_2(s, a) 
&= -0.05 \cdot \Bigl((q_1 - \pi)^2 + q_2^2\Bigr) \\
&\quad - \biggl[
    0.02 \cdot \Bigl(\dot{q}_1^2 + \dot{q}_2^2\Bigr)
    + 0.25 \cdot \Bigl(a^2 + 2 \lvert a\rvert\Bigr) \\
&\qquad\quad
    + 0.02 \cdot \biggl\lvert \frac{a - a_{\text{prev}}}{dt} \biggr\rvert
    + 0.05 \cdot (\dot{q_i}\cdot a)\,\beta
\biggr],
\end{aligned}
\end{equation}
with the state $s$ being the angles of both joints $q_1$ and $q_2$, and their
derivatives. A regularization term (squared brackets) is subtracted which
includes the angular velocities, the agent's action $a$ and the previous action
$a_{\text{prev}}$.
For pendubot, $\beta$ was set to $\beta = 0.1$ and $\dot{q}_i =
\dot{q}_1$ and for acrobot $\beta = 0.025$ and $\dot{q}_i = \dot{q}_2$.
In this application it turned out that providing the agent with negative values
(punishments) instead of positive values (rewards) leads to better performance
and higher robustness.
Having the highest possible value of the reward function at $0$ makes learning
the Q-function much more stable than having an optimal value that is positive,
since a successful swing-up policy does not suddenly change the Q-values of all
previous states and thus prevents policy degradation after learning successful
swing-ups.  

Successful swing-up attempts on the real system using a policy that was purely
trained in simulation require an accurate identification of the physical
parameters of the double pendulum model. Differential Evolution~\cite{c22},
which is a gradient-free optimization algorithm, was used to optimize for the
physical parameters of the simulated double pendulum $\theta_m$ to minimize the
sim-to-real gap. The cost function for the optimization was chosen to be an
importance-weighted squared error between simulated and real trajectories:
\begin{equation}
\label{cost_fcn_history_sac}
\begin{aligned}
J(\theta_m) 
&= \sum_{i=1}^{N_{\text{traj}}}
    \sum_{t=1}^{T_i}
    \sum_{j=1}^{m}
    \Bigl( 1 - \frac{0.5\,(t - 1)}{T_i - 1} \Bigr) \\
&\quad \times
    \Bigl( x_{\text{sim}, j}^{(i)}[t; \theta_m]
           - x_{\text{real}, j}^{(i)}[t] \Bigr)^2.
\end{aligned}
\end{equation}
Here $m = 4$ the number of state indices, $T_i$ is the time steps in a
trajectory, and $N_{\text{traj}}$ the number of trajectories. 

Data from the real system was collected using swing-up policies that were
trained in simulation for acrobot and pendubot. The torque time series from
these trajectories is then used to create trajectories on the simulated system,
resulting in pairs of trajectories for a single torque curve,
$x_{1:T}^{\text{real}}$ and  $x_{1:T}^{\text{sim}}$. The first round brackets in
(\ref{cost_fcn_history_sac}) denote a time-dependent weight. The
weighting decreases linearly over the trimmed trajectory from $1$ to $0.5$, so
that later points in the trajectory with a larger error accumulation have less
impact. Since a solution for the optimization problem becomes increasingly hard
to find with a longer trajectory, only the first $1.5$ seconds of each
trajectory are considered in the optimization.

For the training process, a multi-environment strategy, incorporating multiple
solutions of the previous system identification process, prevented overfitting
on a single environment.

%\subsection{Model-based Optimal Control: TVLQR  (500 words, 1 fig max)}

\section{Results}

% put here for figure placement
\begin{figure*}[t]
    \centering
    \input{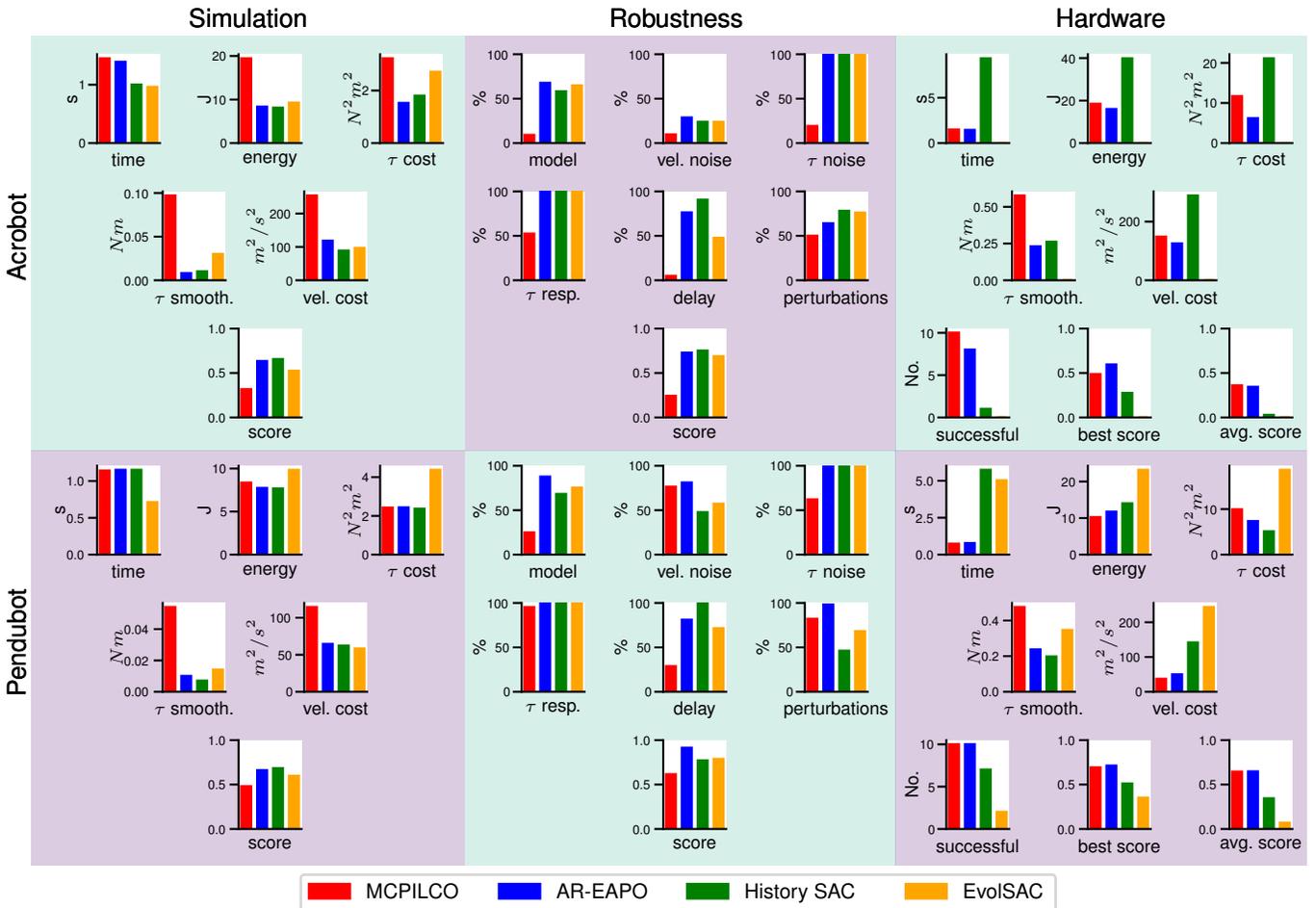}
    \caption{All scores of the final round controllers. The simulation and
    robustness columns are computed from simulation data while the hardware
    score is evaluated from experiments on the real system. Note that for the
    simulation and hardware criteria lower values are better while for the
    success scores in the robustness tests higher values are better. For the
    final scores 1.0 is the best achievable rating.}
    \label{fig:results_barplot}
\end{figure*}

The computed scoring criteria and final scores for all controllers are
visualized in Fig. \ref{fig:results_barplot}. Data, figures and videos of the
individual attempts can be found on the online
leaderboards\footnote{\href{https://dfki-ric-underactuated-lab.github.io/real_ai_gym_leaderboard/}{https://dfki-ric-underactuated-lab.github.io/real\_ai\_gym\_leaderboard/}}.
The accompanying video of this paper shows the swing-ups in simulation and on
the real hardware as well as the controllers robustness to disturbances induced
by hitting the pendulum with a stick.

\subsection{Acrobot}
\begin{figure}[t]
    \centering
    \includegraphics[width=\linewidth]{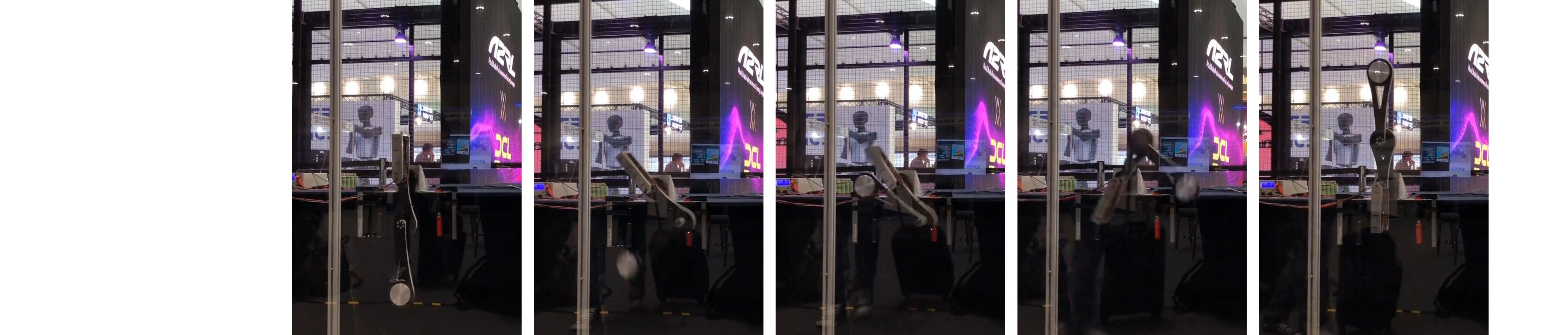}
    \hfill
    \input{images/timeseries_mcpilco_acrobot_real_swingup.pgf}
    \caption{Swing-up trajectory with MC-PILCO on the real acrobot system. From
    top to bottom, the plot shows the evolution of the joint angles, velocities,
    motor torques and disturbances.}
    \label{fig:results_mcpilco_acrobot}
\end{figure}

In the simulation phase HistorySAC and AR-EAPO showed the best performances
(0.66 and 0.63) with a fast swing-up below $1\,\text{s}$, little and smooth
torque usage and low velocities. They also showed good robustness results (0.75
and 0.73) especially concerning robustness to torque noise, torque
responsiveness and delay. Like all other controllers, they were most sensitive to
velocity noise. EvolSAC showed convincing results as well (perf. 0.52, robust.
0.69) only slightly behind HistorySAC and AR-EAPO. MC-PILCO was the least
efficient controller of these four (perf. 0.31), with a higher energy usage and
a less smooth torque signal. In the robustness score MC-PILCO was very sensitive
in all criteria except for torque responsiveness and perturbations (robust.
0.24). In the hardware phase, the model based algorithm MC-PILCO was trained
based on recorded data from the real system which resulted in a policy which
captured knowledge about effects which are not considered in the simple
mathematical model. The policy proved to be very robust to the unknown external
perturbations and MC-PILCO achieved 10/10 successful swing-ups (hardware avg.
score 0.36). The swing-up trajectory of the highest scoring attempt is shown in
Fig. \ref{fig:results_mcpilco_acrobot}.
AR-EAPO achieved even better scores in all five criteria but only managed 8/10
successful swing-ups resulting in a better best trial but a very close second
place in the average score (avg. score 0.34). HistorySAC achieved only
1/10 swing-ups in the final evaluation (avg. score 0.03) and for EvolSAC the
simulation-reality gap was too significant to swing-up the real acrobot system
(avg. score 0.00).

\subsection{Pendubot}
\begin{figure}[t]
    \centering
    \includegraphics[width=\linewidth]{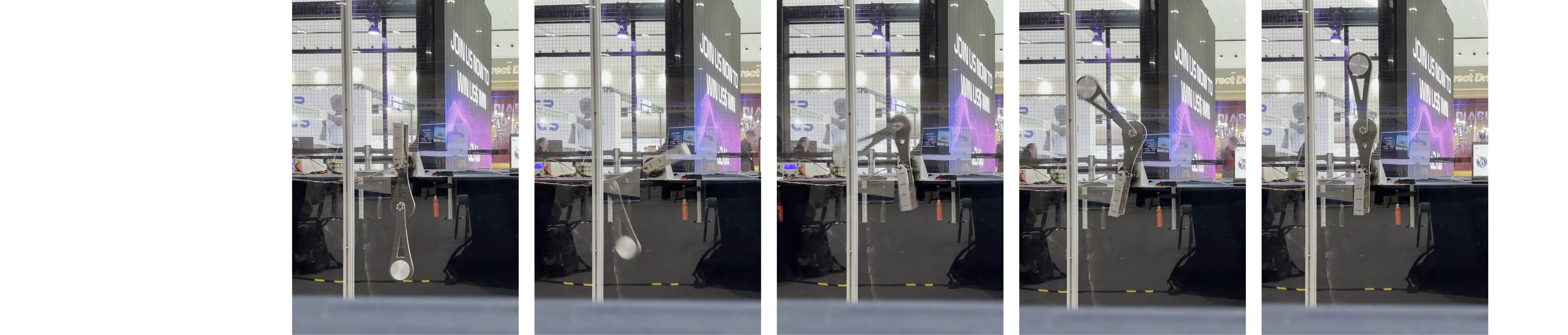}
    \hfill
    \input{images/timeseries_areapo_pendubot_real_swingup.pgf}
    \caption{Swing-up trajectory with AR-EAPO on the real pendubot system. From
    top to bottom, the plot shows the evolution of the joint angles, velocities,
    motor torques and disturbances.}
    \label{fig:results_areapo_pendubot}
\end{figure}

On the pendubot, all four controllers showed similar performances in swing-up
time and energy usage. EvolSAC is the only one with a significantly higher
torque cost and MC-PILCO the only one with a significantly higher torque
smoothness value and velocity cost. This results in similar performance scores
(HistorySAC: 0.68, AR-EAPO: 0.66, EvolSAC: 0.60, MC-PILCO: 0.48). In the
robustness metrics, MC-PILCO again was sensitive to modeling errors and delay.
AR-EAPO was quite robust to modeling errors and also very robust (100\% success)
to random perturbations. This resulted in good robustness scores for all
controllers (AR-EAPO: 0.91, EvolSAC: 0.79, HistorySAC: 0.77, MC-PILCO: 0.61) and
AR-EAPO winning this category. On the real hardware, MC-PILCO and AR-EAPO showed
very convincing performances with 10/10 successful swing-ups and good scores in
the individual criteria. Here, AR-EAPO took the win with a very slight margin
(AR-EAPO: 0.65, MC-PILCO: 0.64). The best AR-EAPO trial is shown in Fig.
\ref{fig:results_areapo_pendubot}. HistorySAC also showed respectable results
with 7/10 swing-ups (avg. score: 0.34). The EvolSAC team was not able to attend
the conference and thus did not fine-tune the controller for the onsite setup,
but EvolSAC was still successful 2/10 times (avg. score: 0.07).

\section{Conclusion}

The '2nd AI Olympics with RealAIGym' competition at the IROS 2024 conference
gives interesting insights into the usability of state of the art RL algorithms
on a real robotic system. The fact that all four teams which advanced to the
final round submitted RL controllers reflects that RL is a very active research
field in robotics. The model based RL method MC-PILCO achieved very good results
with a high robustness by learning very sample efficiently from data recorded
from the real system and won the acrobot competition track. The other method
which stood out was AR-EAPO, with good scores on both systems and winning the
pendubot category. % add an explanation for the good AR-EAPO performance
Both methods exhibited a high robustness to unknown perturbations and even
recovered most of the time after being pushed far away from their nominal path
with a stick, proving to be promising candidates for finding a global swing-up
and balance policy for the underactuated double pendulum systems.

We made sure to make all results online available and hope that this competition
inspires further research and thorough comparisons of control methods for
dynamic behaviors on underactuated robots with high robustness.

%%%%%%%%%%%%%%%%%%%%%%%%%%%%%%%%%%%%%%%%%%%%%%%%%%%%%%%%%%%%%%%%%%%%%%%%%%%%%%%%

%%%%%%%%%%%%%%%%%%%%%%%%%%%%%%%%%%%%%%%%%%%%%%%%%%%%%%%%%%%%%%%%%%%%%%%%%%%%%%%%

%%%%%%%%%%%%%%%%%%%%%%%%%%%%%%%%%%%%%%%%%%%%%%%%%%%%%%%%%%%%%%%%%%%%%%%%%%%%%%%%
%\section*{APPENDIX}

\section*{ACKNOWLEDGMENT}

The competition organizers  at DFKI-RIC (FW, SV, DM, FK, SK) acknowledge the
support of M-RoCK project funded by the German Aerospace Center (DLR) with
federal funds (Grant Number: FKZ 01IW21002) from the Federal Ministry of
Education and Research (BMBF) and is additionally supported with project funds
from the federal state of Bremen for setting up the Underactuated Robotics Lab
(Grant Number: 201-342-04-1/2023-4-1).
TV, BB, and JP acknowledge the grant ``Einrichtung eines Labors des Deutschen
Forschungszentrum für Künstliche Intelligenz (DFKI) an der Technischen
Universität Darmstadt'' of the Hessisches Ministerium für Wissenschaft und
Kunst.
This research was supported by Research Clusters “The Adaptive Mind” and “Third
Wave of AI”, funded by the Excellence Program of the Hessian Ministry of Higher
Education, Science, Research and the Arts.
Parts of the calculations for this research were conducted on the Lichtenberg
high-performance computer of the TU Darmstadt.
Alberto Dalla Libera and Giulio Giacomuzzo were supported by PNRR research
activities of the consortium iNEST (Interconnected North-Est Innovation
Ecosystem) funded by the European Union Next GenerationEU (Piano Nazionale di
Ripresa e Resilienza (PNRR) – Missione 4 Componente 2, Investimento 1.5 – D.D.
1058  23/06/2022, ECS\_00000043). This manuscript reflects only the Authors’
views and opinions, neither the European Union nor the European Commission can
be considered responsible for them.
TV and BB also thank Daniel Palenicek and Tim Schneider for their support while
performing the experiments on the IAS group's computing cluster at TU Darmstadt.

%%%%%%%%%%%%%%%%%%%%%%%%%%%%%%%%%%%%%%%%%%%%%%%%%%%%%%%%%%%%%%%%%%%%%%%%%%%%%%%%

\addtolength{\textheight}{-14.8cm}   % This command serves to balance the column lengths

\bibliographystyle{IEEEtran}
\bibliography{references}

\end{document}